\documentclass{llncs}

\usepackage{makeidx}  

\usepackage{comment}
\usepackage{array}
\newcolumntype{C}[1]{>{\centering\let\newline\\\arraybackslash\hspace{0pt}}m{#1}}
\usepackage{gensymb} 	
\usepackage{graphicx}
\usepackage{subfigure}
\usepackage[dvipsnames]{xcolor} 
\usepackage[margin=4cm]{geometry}

\usepackage{soul}  
 
\begin{document}

\title{Deep driven fMRI decoding of visual categories}


\author{Michele Svanera\inst{1} \and Sergio Benini \inst{1} \and 
Gal Raz\inst{2} \and \\ Talma Hendler\inst{2} \and 
Rainer Goebel\inst{3} \and Giancarlo Valente\inst{3}}

\institute{Department of Information Engineering, University of Brescia, Italy
\and
The Tel Aviv Center for Brain Functions, Tel Aviv Sourasky Medical Center, Israel
\and
Department of Cognitive Neuroscience, Maastricht University, The Netherlands}

\authorrunning{Svanera et al.} 

\maketitle

\begin{abstract}

Deep neural networks have been developed drawing inspiration from the brain visual pathway, implementing an end-to-end approach: from image data to video object classes.
However building an fMRI decoder with the typical structure of Convolutional Neural Network (CNN), i.e. learning multiple level of representations, seems impractical due to lack of brain data.
As a possible solution, this work presents the first hybrid fMRI and learnt deep features decoding approach: collected fMRI and deep learnt representations of video object classes are linked together by means of Kernel Canonical Correlation Analysis.
In decoding, this allows exploiting the discriminatory power of CNN by relating the fMRI representation to the last layer of CNN (\textit{fc7}).
We show the effectiveness of embedding fMRI data onto a subspace related to deep features in distinguishing two semantic visual categories based solely on brain imaging data.

\end{abstract}

\section{Introduction}

Understanding brain mechanisms associated with sensory perception is a long-standing goal of Cognitive Neuroscience. Using non-invasive techniques, such as fMRI or EEG, researchers have the possibility to validate computational models by relating stimulus features and brain activity. 
Several approaches have been proposed in the literature to establish this link, using both generative and discriminative models, in combination with machine learning techniques.
These techniques have become a common tool for the analysis of neuroimaging data, due to their ability to identify robust associations between high dimensional imaging datasets and subject's perception.


Among the possible taxonomies of the current literature, the distinction between \emph{encoding} and \emph{decoding} models (see \cite{NKN11}), when applied to imaging data, is of particular interest in the context of this work.
In \emph{decoding} models the aim is to learn a distributed model capable of predicting, based on the associated measurements, a categorical or a continuous label associated with a subject's perception. 
This approach, often referred to as Multi-Voxel Pattern Analysis (MVPA), uses classification and identifies a discriminating brain pattern that can be used to predict the category of new, unseen stimuli. 
When the labels are instead continuous, multivariate regression approaches are used to create a link between a target and a multivariate model. 
An example is the PBAIC brain reading competition \cite{VME11}, where the participants had to predict, based on fMRI data associated with a virtual reality (VR) world experiment, specific continuous features, derived from automatic annotation of the VR experience and from eye-tracking data. 
Another application of these approaches is in \cite{RSC17}, where a robust multivariate model, based on data acquired when subjects watch excerpts from movie, is used to predict low-level and semantic features across subjects and movies. 
In the vast majority of decoding study, the problem is to link an $n$-dimensional dataset to a $1$-dimensional target (associated with subjects' perception). When multiple targets are available, standard approaches consider one task at a time.  
\emph{Encoding} models, on the other hand, take the complementary approach and consider the association between $n$-dimensional features related to subject's perception, and each brain voxel \cite{KNP08}. 

Going beyond $n$ to $1$ and $1$ to $n$, unsupervised models have been proposed to link multivariate representations in the data and in the subject experience dimensions. Canonical Correlation Analysis (CCA) is particularly suited in this respect, as it allows projecting one dataset onto another by means of linear mapping, which can be further used for categorical discrimination and brain models interpretations. In \cite{HMB07} kernel CCA has been used to create associations between visual features of static pictures and associated brain patterns. In \cite{BSB11} the authors consider six descriptive features extracted from a movie and linked them with fMRI data with semi-supervised Kernel CCA, exploring different variants and regularization types. 
fMRI-derived responses have been used, through CCA, as integrated semantic features in addition to low-level features for classification of videos categories \cite{HDL10} or music/speech categories \cite{JZH12,TCT15}.


When dealing with visual stimuli, the brain imaging community is making more and more use of deep neural networks, since they provide high capability and flexibility in image and video description. 
On the other hand, the deep neural network community has always been inspired by the brain mechanisms while developing new methods. The two communities therefore share now many common research questions. For example, how the brain transforms the low-level information (colors, shapes, etc.) into a certain semantic concept (person, car, etc.) is an important research topic for both the communities.
In \cite{GvG15} the authors use the same dataset as in \cite{KNP08} representing the stimuli with more abstract features, derived from deep neural networks. 
The same approach has been used in \cite{ASM14}, which introduces new class of encoding models that can predict human brain activity directly from low-level visual input (i.e., pixels) with ConvNet \cite{KSH12}.
In \cite{CHY14} a direct performance comparison is performed between deep neural networks and IT neurons population recorded in a monkey.
The work in \cite{KK14} exploits a representational similarity analysis in which the two representations (fMRI and deep features) are characterized by their pairwise stimulus correlation matrix. 
For a given set of stimuli, this matrix describes how far apart appear two representations of the same stimuli.
Related to representations comparison, the work in \cite{CKP16} compared fMRI representations with artificial deep neural network representations tuned to the statistics of real-world images.
A good overview of these technique could be found in \cite{YD16}.

Representations for video object classes usually present in the common movies (i.e. person, dog, car, etc.), are obtained analysing frames with \textit{faster R-CNN} method \cite{RGH16} and extracting the last fully connected layer as proper feature, as originally proposed in \cite{DJV14}.
These detection networks with ``attention'' mechanism are based on a deep convolutional framework, called Region Proposal Networks (RPNs), which produces object proposals using the output of convolutional layers.
Despite some recent works on video description are very promising for correlating also in temporal domain, the proposed method adopts faster R-CNN which is the basis of current state-of-the-art for object class detection \cite{HZR15}.

\subsection{Paper aims and contributions}

The principal aim of this work is to build a decoder model with a $n$ to $m$ approach, linking fMRI data taken while watching natural movies and deep neural network video descriptions extracted with \textit{faster R-CNN} method.  
The obtained model is able to reconstruct, using fMRI data, the deep features and exploit their discrimination ability. 
This goal is achieved using Canonical Correlation Analysis (CCA) \cite{B36} and its kernel version (kCCA), which relates whole-brain fMRI data and video descriptions, finding the projections of these two sets in two new spaces that maximise the linear correlation between them.
%

To validate the proposed method, different tests are conducted across multiple subjects and various movies. 
Preliminary results for a classification task are provided, showing the ability of the method to embed brain data into a space more suitable for the discrimination task of distinguishing in frames the presence of \textit{faces} and \textit{full human figures} (ground truth is manually annotated for every TR).
To the best of our knowledge, this is the first work that try to combine fMRI data and learnt deep features to perform a decoding scheme able to distinguish between two video object categories from the observed fMRI response.

\section{Proposed method}
\label{sec:proposed_method}

\subsection{fMRI acquisition and preparation}

Data are collected from several independent samples of healthy volunteers with at least 12 years of education using a $3$ Tesla GE Signa Excite scanner.
Data for all movies are part of a larger dataset collected for projects examining hypotheses unrelated to this study.
In Table~\ref{table:movies_and_VTC} relevant information about movies and subjects are reported: title, duration and some subject properties.
%
\vspace{-0.5cm}
\begin{table}[ht]
\caption{Movie dataset} 
\centering 
\begin{tabular}{| p{6.5cm} | C{1.5cm} | C{1.4cm} | C{2cm} | C{1.4cm} |} 
\hline
\textbf{Film title} & \textbf{Duration (mm:ss)} & \textbf{Subjects} &  \textbf{Average $\pm$std age (years)} &   \textbf{Female/ Male}\\ 
\hline
Avenge But One of My Two Eyes (Mograbi, 2005) & 	5:27 & 74	& 	19.51$\pm$1.45	&   0/74\\
Sophie's Choice (Pakula, 1982) &  10:00 & 44  &  26.73$\pm$4.69 &	25/19\\
Stepmom (Columbus, 1998) & 	8:21 & 53	& 	26.75$\pm$4.86 &	21/32 \\
The Ring 2 (Nakata, 2005) &	 8:15 & 27 & 	26.41$\pm$4.12 &	11/16\\
The X-Files, episode ``Home'' (Manners, 1996)	&  5:00 & 36 & 	23.70$\pm$1.23	& 14/22\\
\hline
\end{tabular}
\label{table:movies_and_VTC} 
\end{table}  
\vspace{-0.3cm}
Due to technical problems and exaggerated head motions (1.5 mm and 1.5\degree from the reference point) only stable data are included.
Functional whole-brain scans were performed in interleaved order with a T2*-weighted gradient echo planar imaging pulse sequence (time repetition [TR]/TE = 3,000/35 ms, flip angle=90, pixel size = 1.56 mm, FOV = 200$\times$200 mm, slice thickness = 3 mm, 39 slices per volume). 
Data are pre-processed and registered to standardised anatomical images via Brainvoyager QX version 2.4 (Brain Innovations, Maastricht, Netherlands). 
Data are high pass filtered at 0.008 Hz and spatially smoothed with a 6 mm FWHM kernel.
For subject clustering and further acquisition details please refer to 
\cite{RTW16}.
We confined the analysis using a gray matter mask based on an ICBM 452 probability map (\cite{ICBM}) thresholded to exclude voxels with probability lower than $80\%$ of being classified as gray matter (thus encompassing both cortical and brain stem regions) obtaining a fMRI data with $\sim$$42000$ voxels.

\subsection{Video object features}

\begin{figure}[h]
\centering
\includegraphics[width=\textwidth]{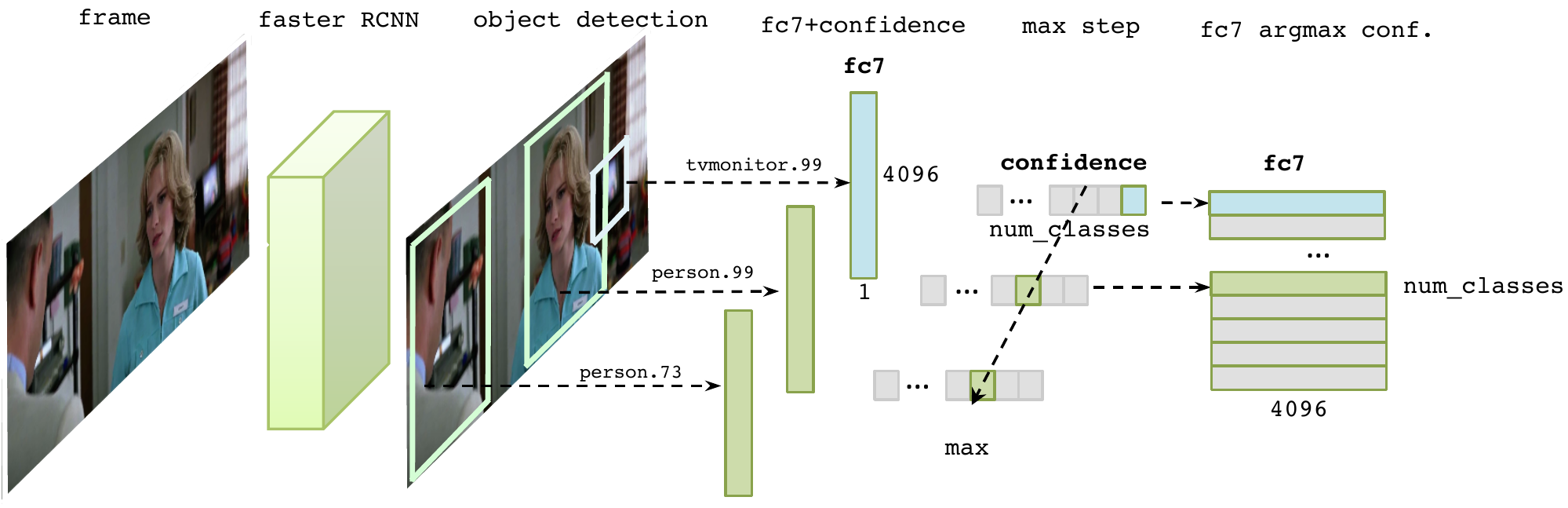}
\caption{Feature extraction procedure for each processed frame in the video (5fps).}
\label{fig:frame_elaboration}
\end{figure}

\begin{figure}[h]
\centering
\includegraphics[width=\textwidth]{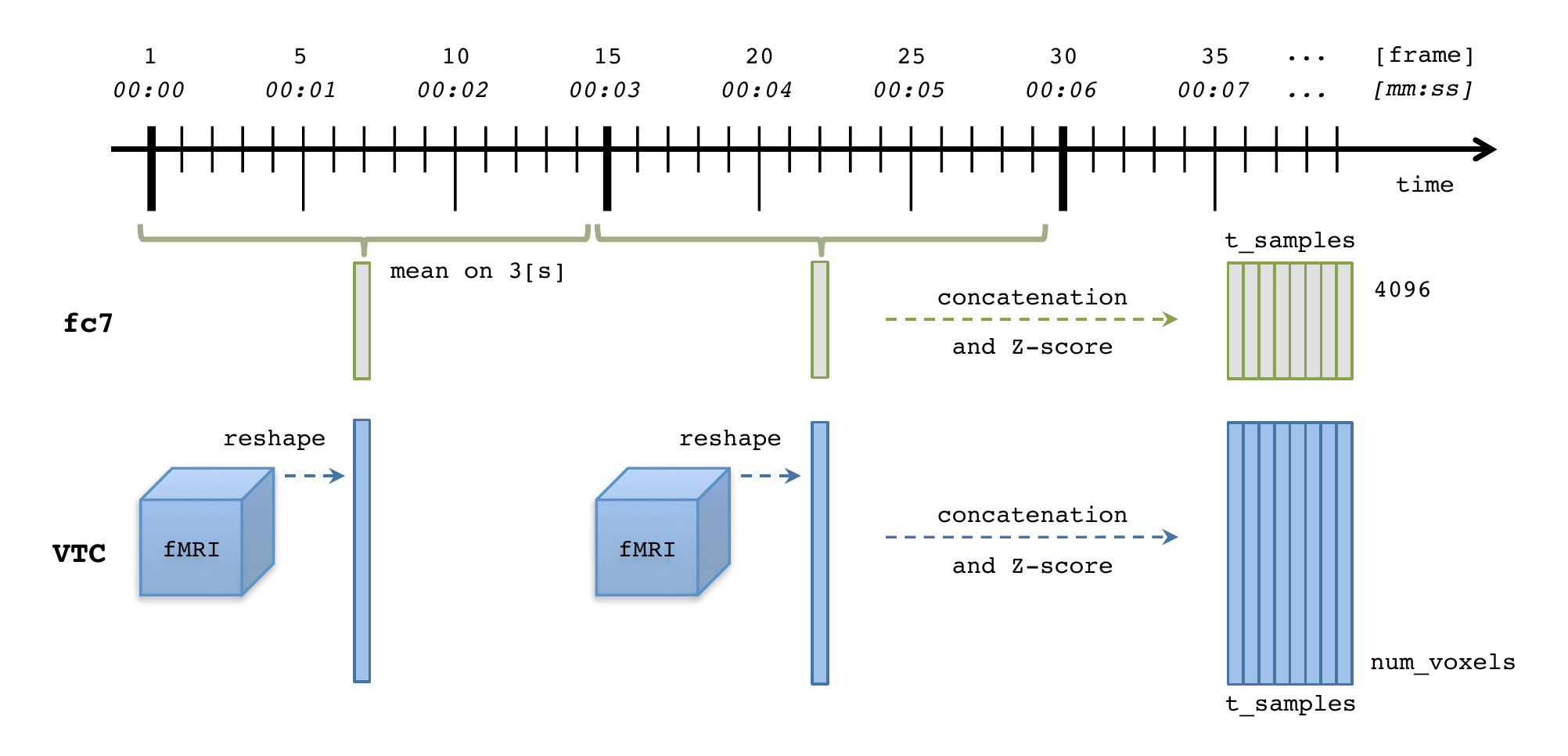}
\caption{Extraction and normalization of $fc7$ features from the video stream (top) and volume time courses (VTC) extraction from fMRI data (bottom) for one movie and one subject. New subjects and movies are concatenated in time dimension.}
\label{fig:rate_align}
\end{figure}

Features are extracted and collected from video frames as described in Figure~\ref{fig:frame_elaboration} and~\ref{fig:rate_align}.
First each processed frame feeds a \textit{faster R-CNN} framework \cite{RGH16}.
Multiple objects, together with their related confidence values and last fully connected layer ($fc7$), are therefore extracted from each processed frame at different scales and aspect ratios.
$fc7$ features are the last fully connected layers before classification (\textit{softmax}) and are considered as highly representative feature of the object class and shape \cite{DJV14}.
Since it is possible to have in one frame multiple detections of the same object class (as in Figure~\ref{fig:frame_elaboration} for the class ``person''), for each class only the $fc7$ layer of the object with maximum confidence is kept.
For this work only ``person'' class is considered, obtaining a $4096$ dimension feature vector from each frame.
Although it may seem a limitation in the analysis, we must be sure that the attention of the subjects while watching movies is placed on the classes in analysis.
Since human figures is central to modern cinematography \cite{BSA16} we can be  confident about results.
In addition, the proposed work can be expanded to different classes without changes in the framework architecture.

The whole procedure is performed at a reasonable frame rate of $5 fps$ on the entire video.
As shown in Figure~\ref{fig:rate_align}, in order to properly align the $fc7$ feature matrix with the VTC data resolution ($3$ $s$), $fc7$ feature vectors are averaged on sets of $15$ frames.
Different subjects and different movies are concatenated in time dimension, keeping the correspondence fMRI and visual stimuli valid: subjects watching equal movie share the same $fc7$ features but different fMRI data.
%

\subsection{Linking method}

We learned multivariate associations between the fMRI data VTC and the deep  features $fc7$ using \textit{Canonical Correlation Analysis} (CCA). 
Originally introduced by Hotelling \cite{B36}, CCA aims at transforming the original datasets by linearly projecting them, using matrices $A$ and $B$, onto new orthogonal matrices $U$ and $V$ whose columns are maximally correlated: the first component (column) of $U$ is highly correlated with the first of $V$, the second of $U$ with the second of $V$ and so on.

\begin{figure}[h]
\centering
\subfigure[\textbf{Training}: $V = VTC \times B$, $U = fc7 \times A$]{
	\includegraphics[width=\textwidth]{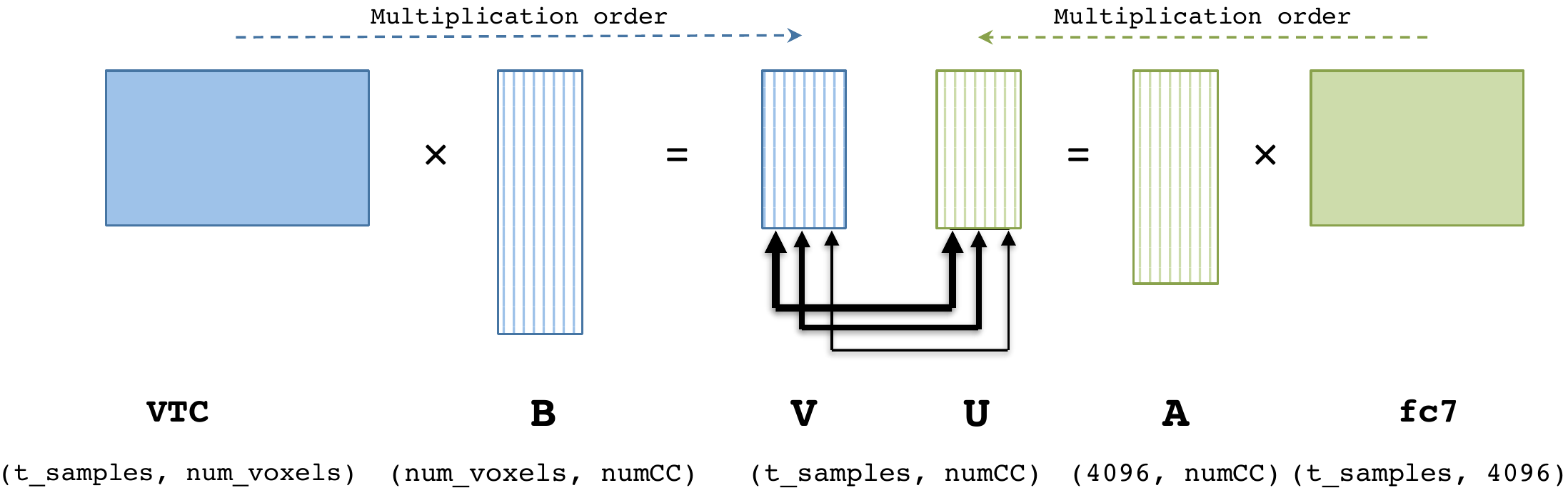}}
\subfigure[\textbf{Testing}: $V = VTC \times B$, $fc7 = U \times A^{+}$]{
	\includegraphics[width=\textwidth]{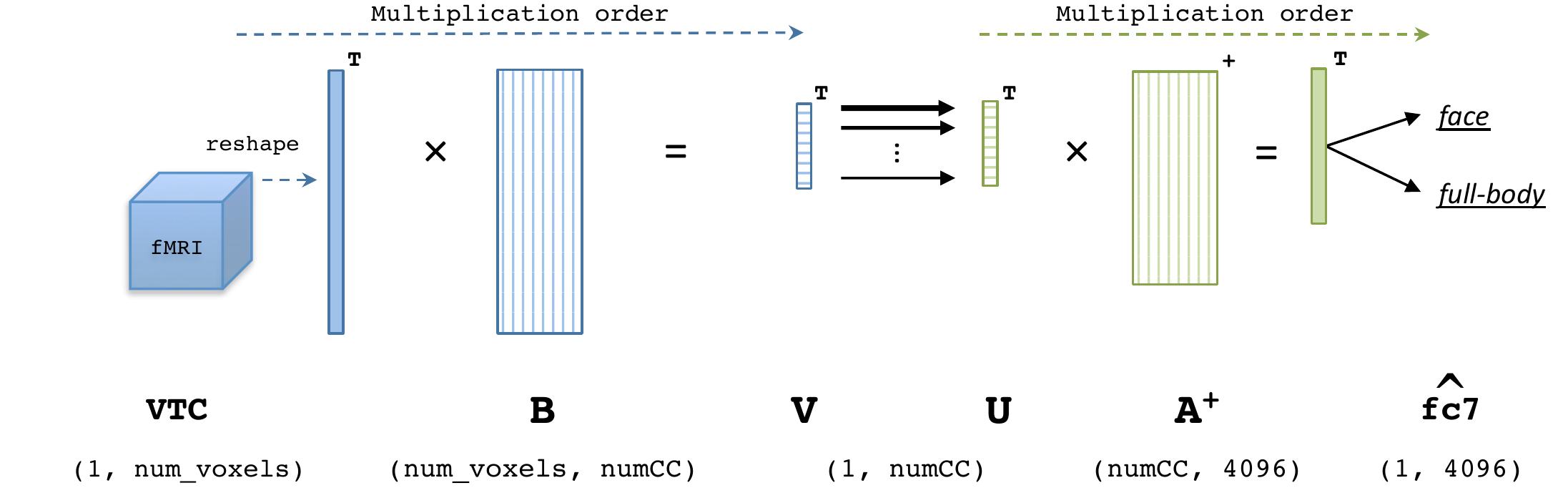}}
\caption{Mapping procedure between video object features ($fc7$) and brain data (VTC) in (a) training (single step) and (b) testing (repeated for every time point). Some matrices are transposed in the figure \textit{exclusively} for graphical purposes: please refer to formulas and displayed matrix dimensions.}
\label{fig:matrices_mapping}
\end{figure}

In training step (Fig.~\ref{fig:matrices_mapping}-a), matrices $U$ and $V$ are obtained from VTC data and $fc7$ features. The correlation between $U$ and $V$ components is validated using new data (different subjects and/or movies) to assess its robustness.
In the testing step (Fig.~\ref{fig:matrices_mapping}-b), the decoding procedure is performed starting from VTC data and obtaining $fc7$ through the matrices $A$ and $B$ previously found.
We show how this scheme can be used to perform a classification task based on the reconstructed $fc7$ matrix.

When the number of time samples is lower than data dimension (i.e. voxels number), the estimation of the canonical components is ill-posed. We therefore used a kernel variant with a linear kernel in combination with a quadratic penalty term (L2-norm), to estimate $A$ and $B$, using the Python module \textit{Pyrcca} \cite{BG15}. We account for the haemodynamic effects in the fMRI signal by introducing a shift between the two datasets (parameter selection in Sec.~\ref{sec:correlation}). Other approaches, such as convolution of deep network features with a canonical haemodynamic model, could be used as well, but in our case they held similar performances and were therefore not explored any further in this study.

\section{Experiments}
\label{sec:experiments}

We examined the data associated with 5 movies described in Table \ref{table:movies_and_VTC}  with $\sim$$230$ VTC scans taken while watching a total of $\sim$$37$ minutes videos.
Experiments are presented as follows: after having set up the framework by parameter tuning in Section~\ref{sec:correlation}, we validate it by generalizing the model to new subjects and/or movies in \ref{ref:GenNewSubjects}, and show classification performance on an exemplary discrimination task in Section~\ref{sec:classification}.

\subsection{Parameter tuning}
\label{sec:correlation}

To tune parameters (regularization and time shift) we used a single movie, \textit{Sophie's Choice}, selecting a random subset of 35 subjects. 
The left out 9 subjects are used for testing in the next steps (see Sec.~\ref{ref:GenNewSubjects} and Sec.~\ref{sec:classification}). 
We further randomly subdivided the 35 training subjects into two disjoint datasets with 30 (training) and 5 (validation) subjects, exploring the effect of temporal shift and regularization on the validation dataset. 

\begin{figure}
\centering
\subfigure[Regularization]
	{\includegraphics[width=0.31\textwidth]{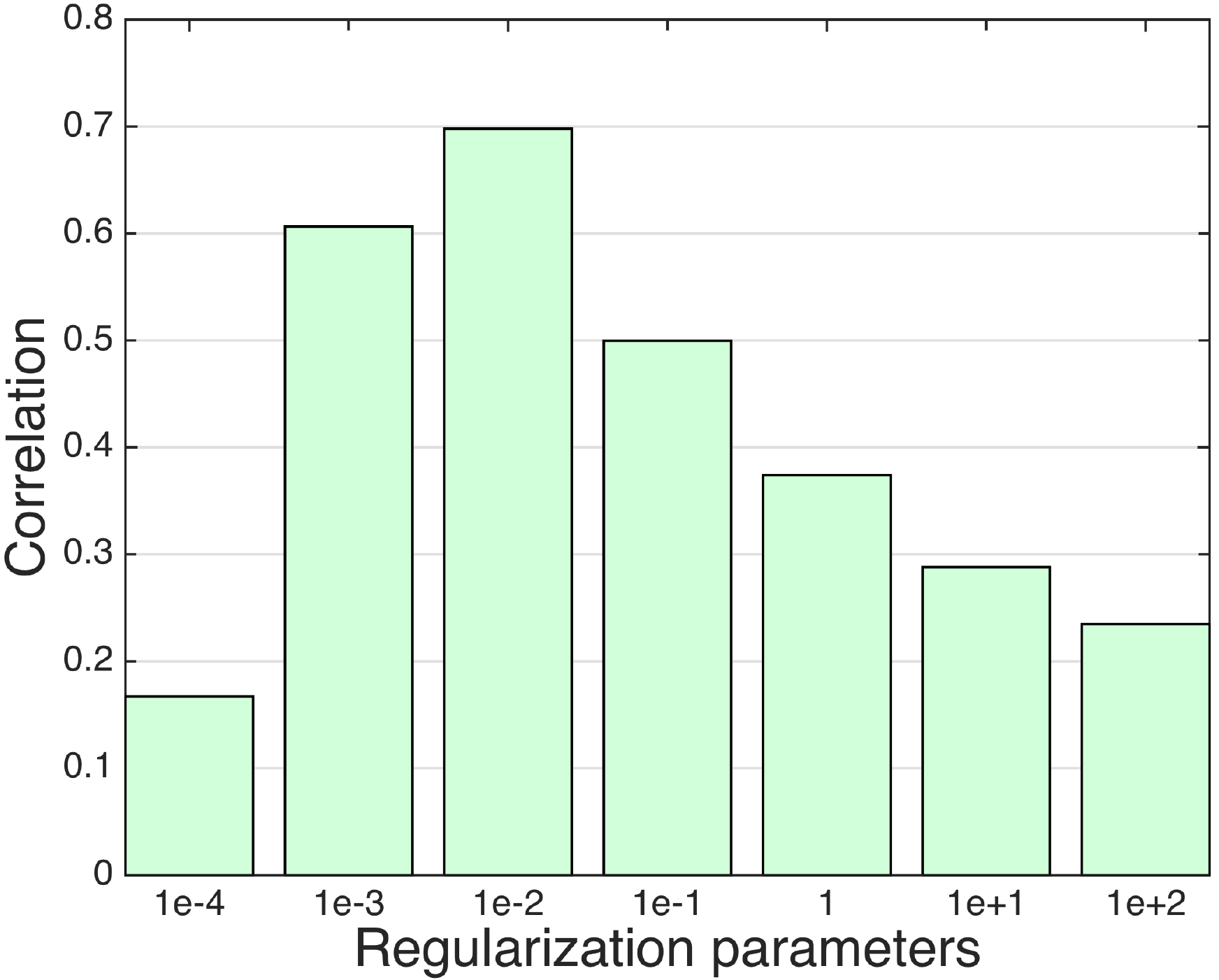}}
\subfigure[Time shift]
	{\includegraphics[width=0.31\textwidth]{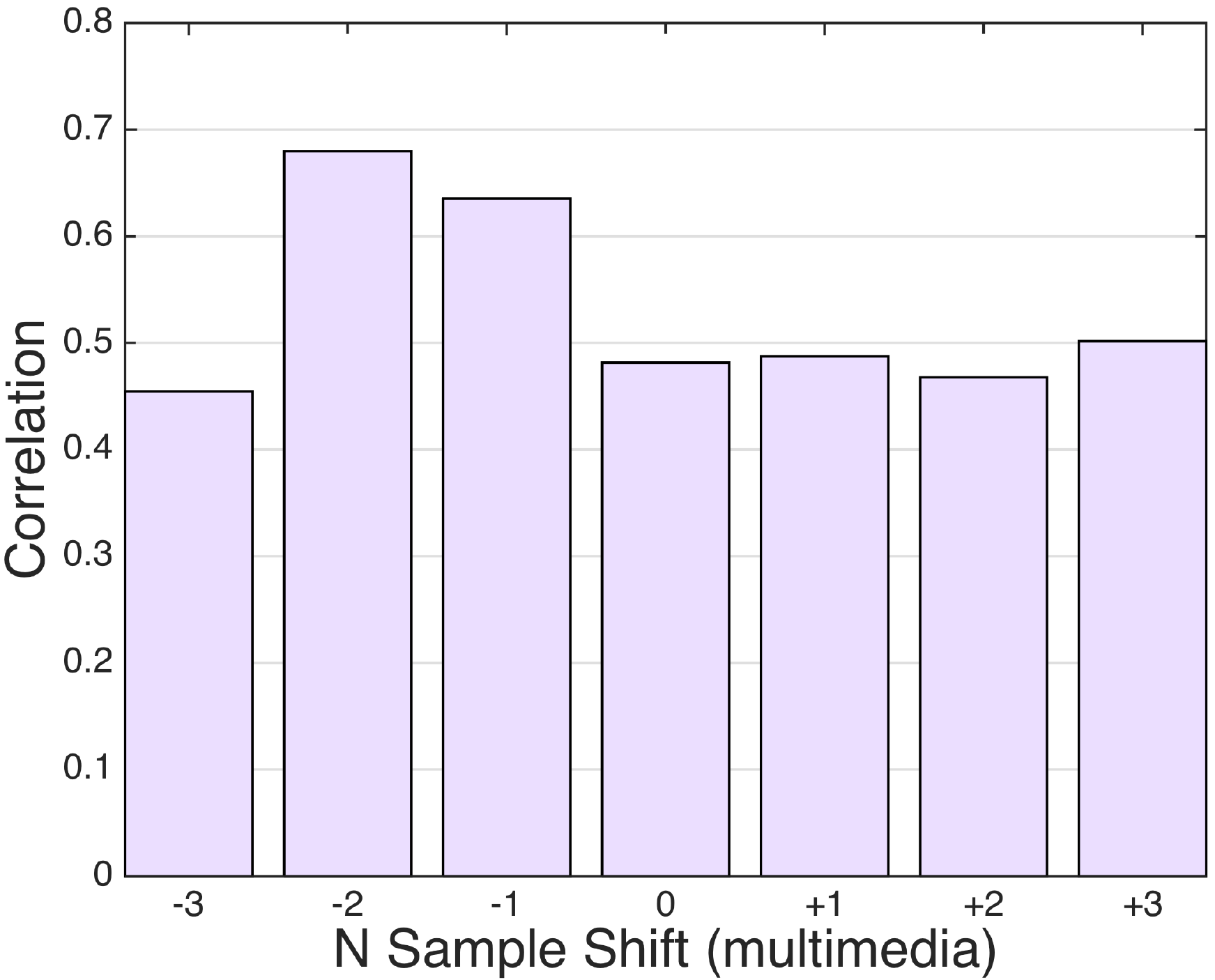}}
\subfigure[Number of components]
	{\includegraphics[width=0.31\textwidth]{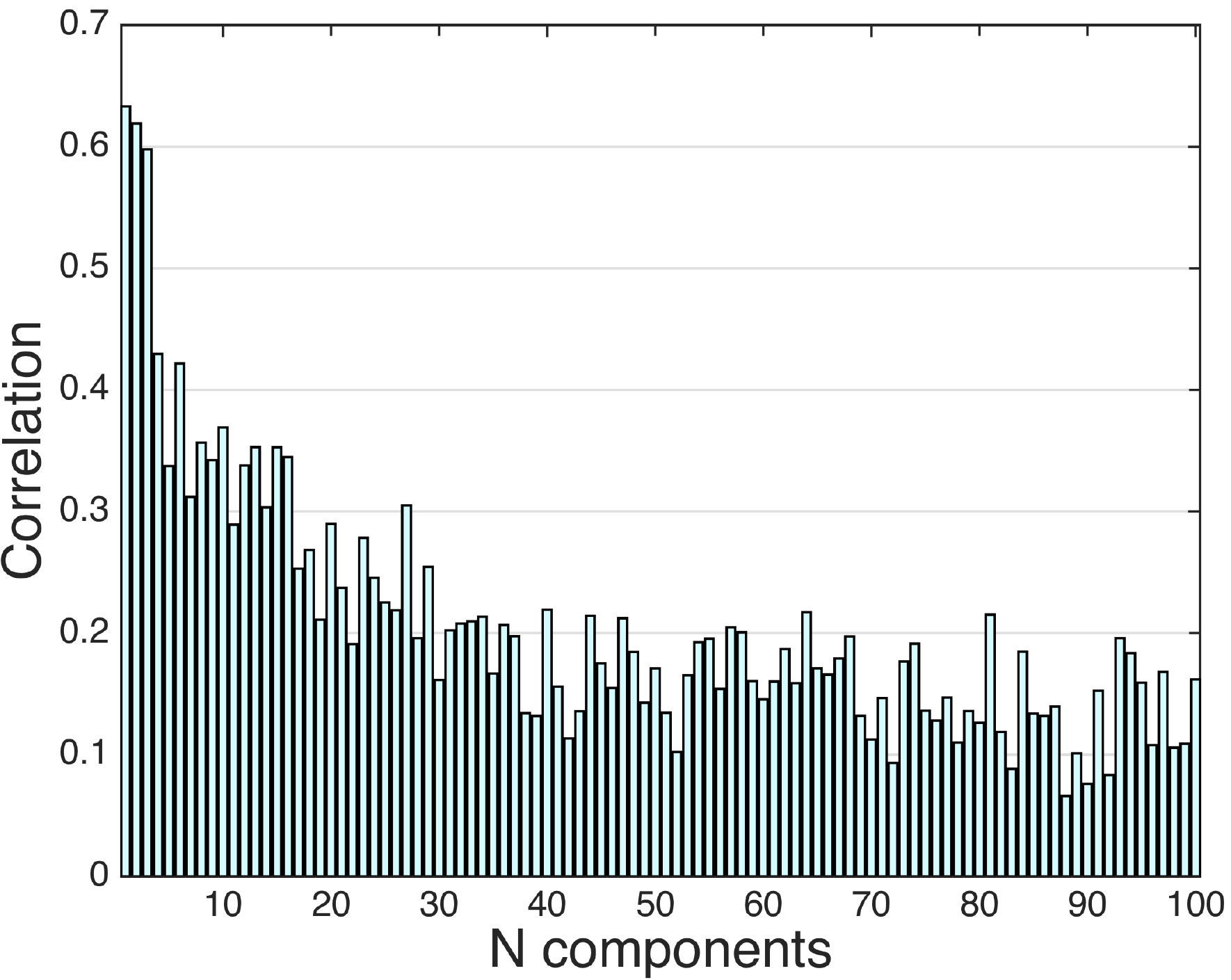}}
\caption{Parameter estimations obtained on training data: (a) regularization ($1^{st}$ component, $ts=-2$), (b) time-shift ($ts$) estimation ($1^{st}$ component, $\lambda=10^{-2}$), (c) correlation between canonical components ($\lambda=10^{-2}$, $ts = -2$).}
\label{fig:k_validation}
\end{figure}

Figure~\ref{fig:k_validation}-a shows the correlation of the first canonical component as a function of the regularization parameter $\lambda$: the value $\lambda=10^{-2}$ ensured the best correlation. 
The second estimated parameter is the time shift between video features and VTC samples.
Different time shifts ($ts$) values were tested: from $ts=-3$ to $+3$ time samples.
Value $ts=-2$ (as in Fig.~\ref{fig:k_validation}-b) returns the highest correlation, in line with what is expected of the hemodynamic response, which peaks 4 to 6 seconds after stimulus onset.
We furthermore explored the correlation values changing the number of canonical components (columns number of $U$ and $V$) in Figure~\ref{fig:k_validation}-c. Large values of correlation were observed for the first three components, with decreasing correlations as the number of components increases.

\begin{figure}
\centering
\includegraphics[width=1\textwidth]{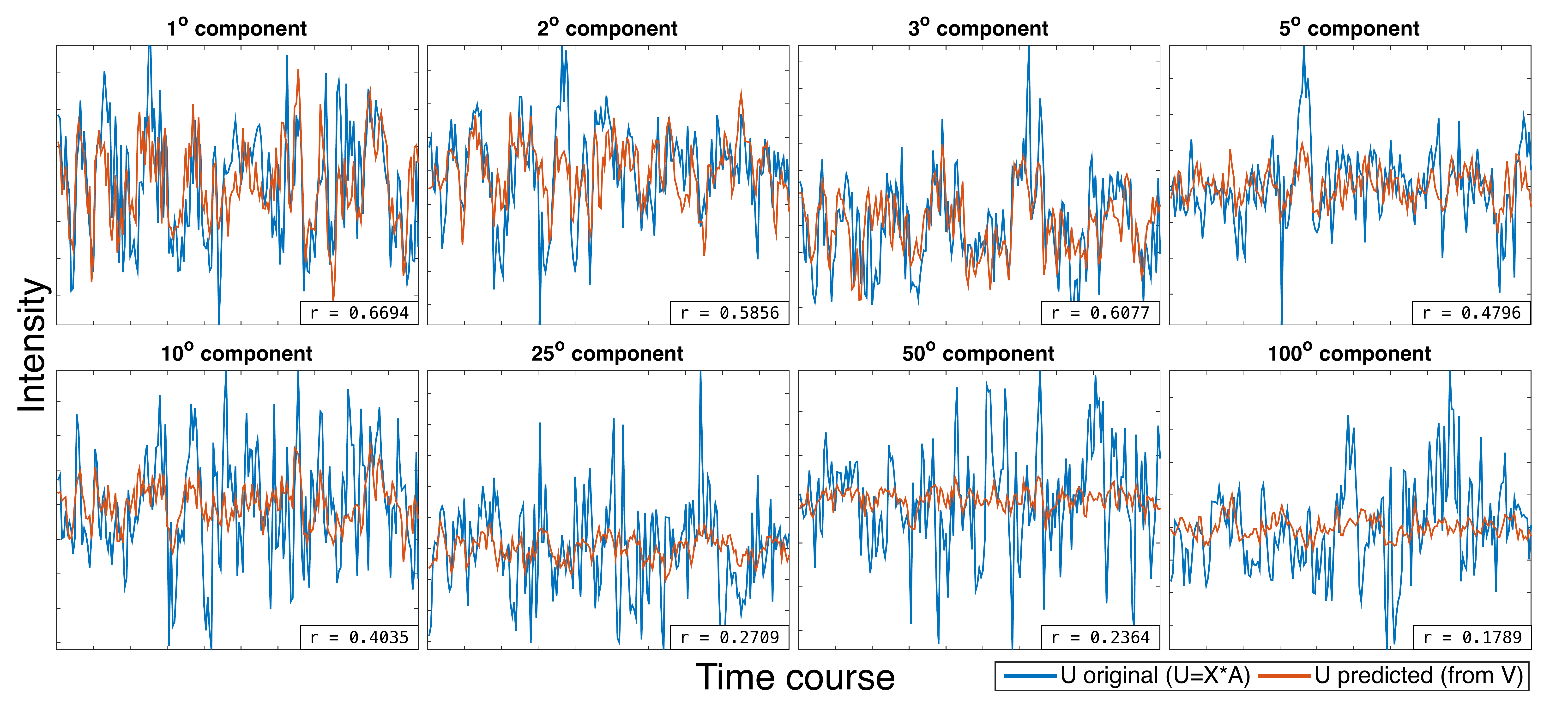}
\caption{Time courses of the original (blue) and regressed (red) signals obtained by one test subject while watching \textit{Sophie's Choice}.}
\label{fig:time_corr_u_U_pred}
\end{figure}

Figure~\ref{fig:time_corr_u_U_pred} shows the time courses of different canonical components of the matrix $U$ (in blue) and their linear reconstruction from the corresponding canonical components of  $V$ (in red) for a single subject while watching the whole movie \textit{Sophie's Choice}.
Due to the linear dependence of $U$ and $V$, a simple linear regression is enough to reconstruct data. 
As indicated by the high correlation value, the first components are able to adequately infer the time course of the matrix $U$ (first row of Fig.~\ref{fig:time_corr_u_U_pred}); conversely, the quality of prediction rapidly decreases after the $10th$ component (second row of Fig.~\ref{fig:time_corr_u_U_pred}). 
It is worth to mention that these plots refer to one subject of the testing set, while the regression weights have been estimated exclusively using the training set. 
%

\subsection{Generalization to new subjects and movies}
\label{ref:GenNewSubjects}

\begin{figure}[h]
\centering
\subfigure[Single movie]
	{\includegraphics[height=0.21\textheight]{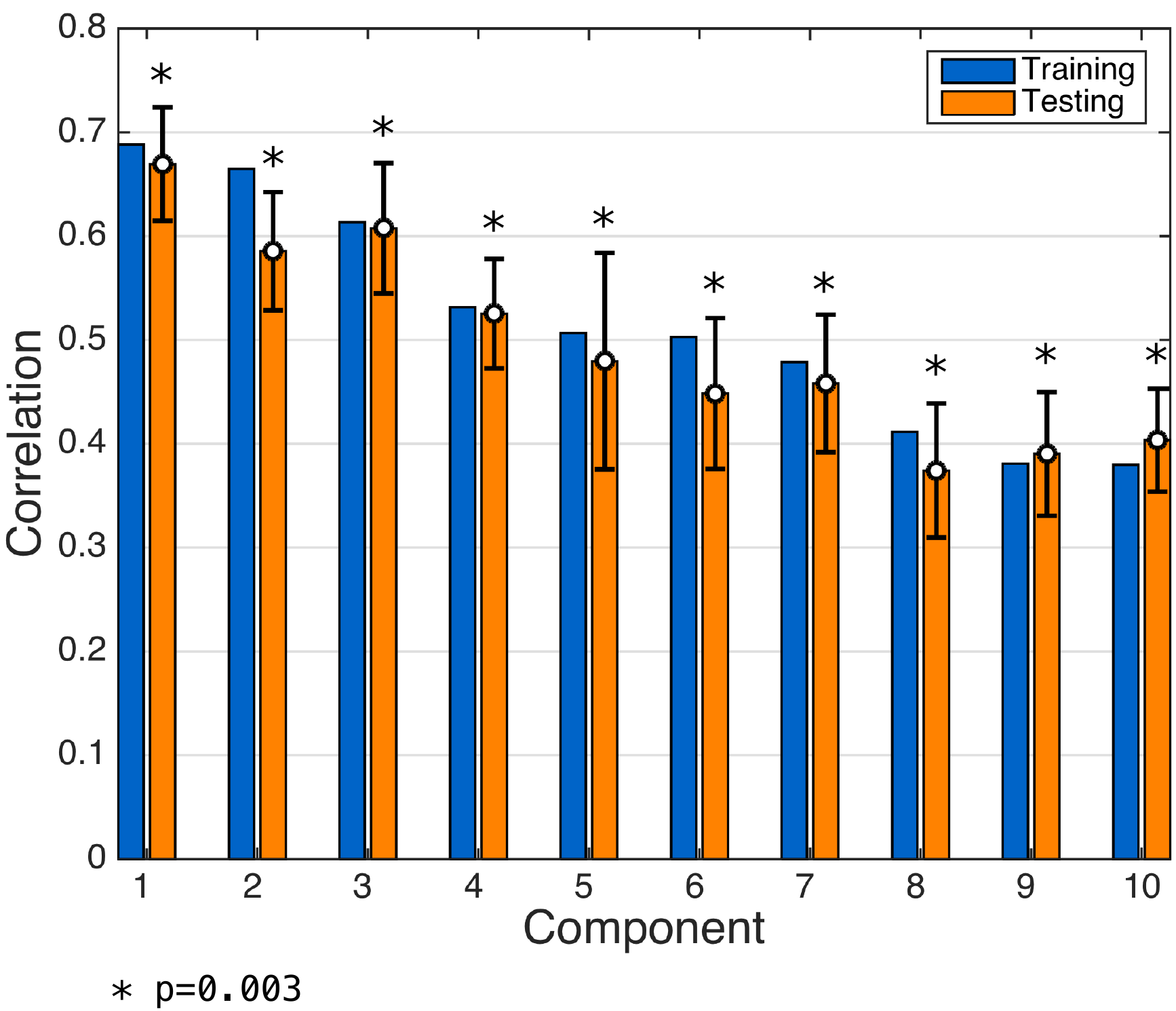}}
\subfigure[Across movies]
	{\includegraphics[height=0.21\textheight]{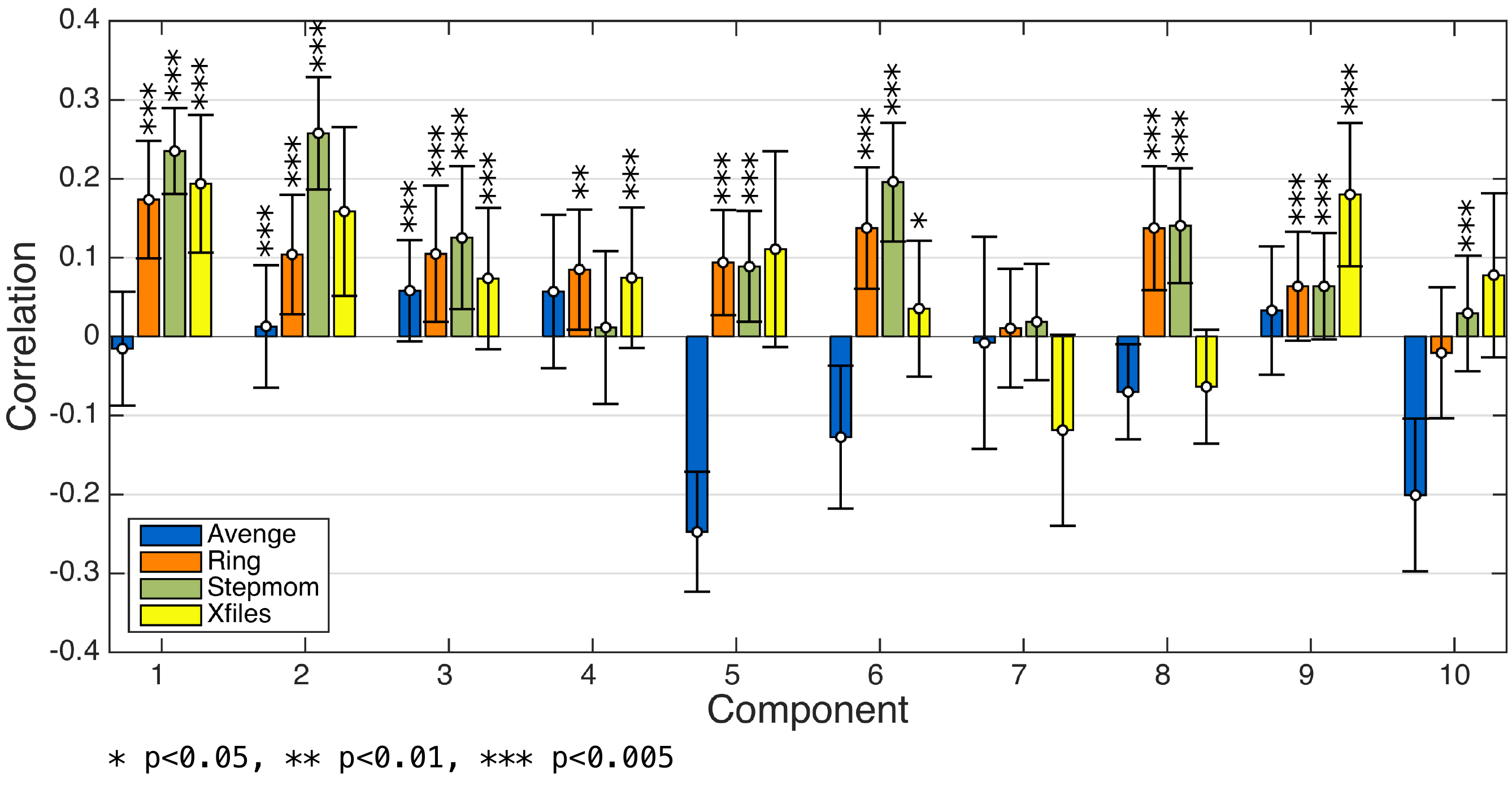}}
\caption{a) Correlation on a single movie (\textit{Sophie's Choice}) on all test subjects (similar values on all movies); b) correlation results across movies and across subjects (training on \textit{Sophie's Choice}).}
\label{fig:corr_across_movies}
\end{figure}

With $\lambda = 10^{-2}$ and a time shift of two samples, we estimated a kCCA model using the 35 training subjects associated with the movie \textit{Sophie's Choice}. 
Figure~\ref{fig:corr_across_movies}-a shows the correlations between the first 10 canonical components on the training and on the left-out, testing dataset (9 subjects). Training and testing feature are permuted scrambling the phase of the Fourier transform respect to original features. The entire training-testing procedure is repeated 300 times on randomly permuted features, and the $p$-values associated with the correlation between canonical projections are calculated. It is worth mentioning that with 300 permutations, the lowest attainable $p$-value, 1/301 (0.003), is obtained when the correlation values observed in the permutations never equal or exceed the correlation obtained on the original data. 

We further explored the robustness of the method by generalizing this model on the remaining  movies. Significance was determined with the phase scrambling permutation process only on testing movies (500 times), leaving unchanged the training set (\textit{Sophie's Choice}). The results are shown in Figure~\ref{fig:corr_across_movies}-b; in this case, three movies (\textit{The Ring 2}, \textit{Stepmom} and \textit{The X-Files}) show significant correlations among the first ten components. 
However the movie \textit{Avenge} does not: one possible explanation for this behaviour can be found in the different stylistic choices adopted by directors in the five movies, for example in the use of the shot scale, light conditions or video motion.
Even if previous correlation results obtained on single movies are good indicators about the soundness of the proposed approach, this stylistic interpretation has to be fully proven in later stages of the work.

\subsection{Classification}
\label{sec:classification}
The last result section shows how the linking between deep neural networks and brain data can be beneficial to subtle classification tasks. 
Since in this work we consider only the $fc7$ features related to the class \textit{person}, we chose, as an example, to discriminate the portion of human figure shot in the video frames, distinguishing between two classes: face only (\textit{face}) or full figure (\textit{full-body}) by conducting three analyses.
Face and full-body ground truth is manually annotated for every TR.
All three analyses are made on single movie \textit{Sophie's Choice}, selecting the 35 training VTCs and the 9 testing VTCs as before.

First, we evaluated classification using whole-brain fMRI data only; a linear SVM classifier was trained using balanced (across classes) training samples selected from the 35 training VTCs  and tested on the 9 testing VTCs.
Given the large dimensions of the fMRI data, the relatively fine-grained difference between the two classes, and the individual differences across subjects, poor performance is expected. 
Second, we classified using $fc7$ features; this could be considered as an upper bound: since these features are inherently capable of discriminating different shapes, excellent results are expected. 
The features in $fc7$ were randomly split into training and testing (75\%-25\%), and a balanced SVM classifier with linear kernel was trained and tested.
Last, we used the proposed link between $fc7$ and VTC and reconstructed the deep neural network features starting from the observed test fMRI data. 
Given a VTC with 1 TR, we follow the pipeline shown in Figure~\ref{fig:matrices_mapping}-b obtaining a $fc7$-like vector. $fc7$ were reconstructed from $V$ using the Moore-Penrose pseudo-inverse of $A$.
We finally learned a balanced linear SVM with 35 training VTCs and testing with the remaining 9.
Different number of canonical components \textit{numCC} were considered.

\begin{figure}
\centering
\subfigure[Accuracy]
	{\includegraphics[width=0.45\textwidth]{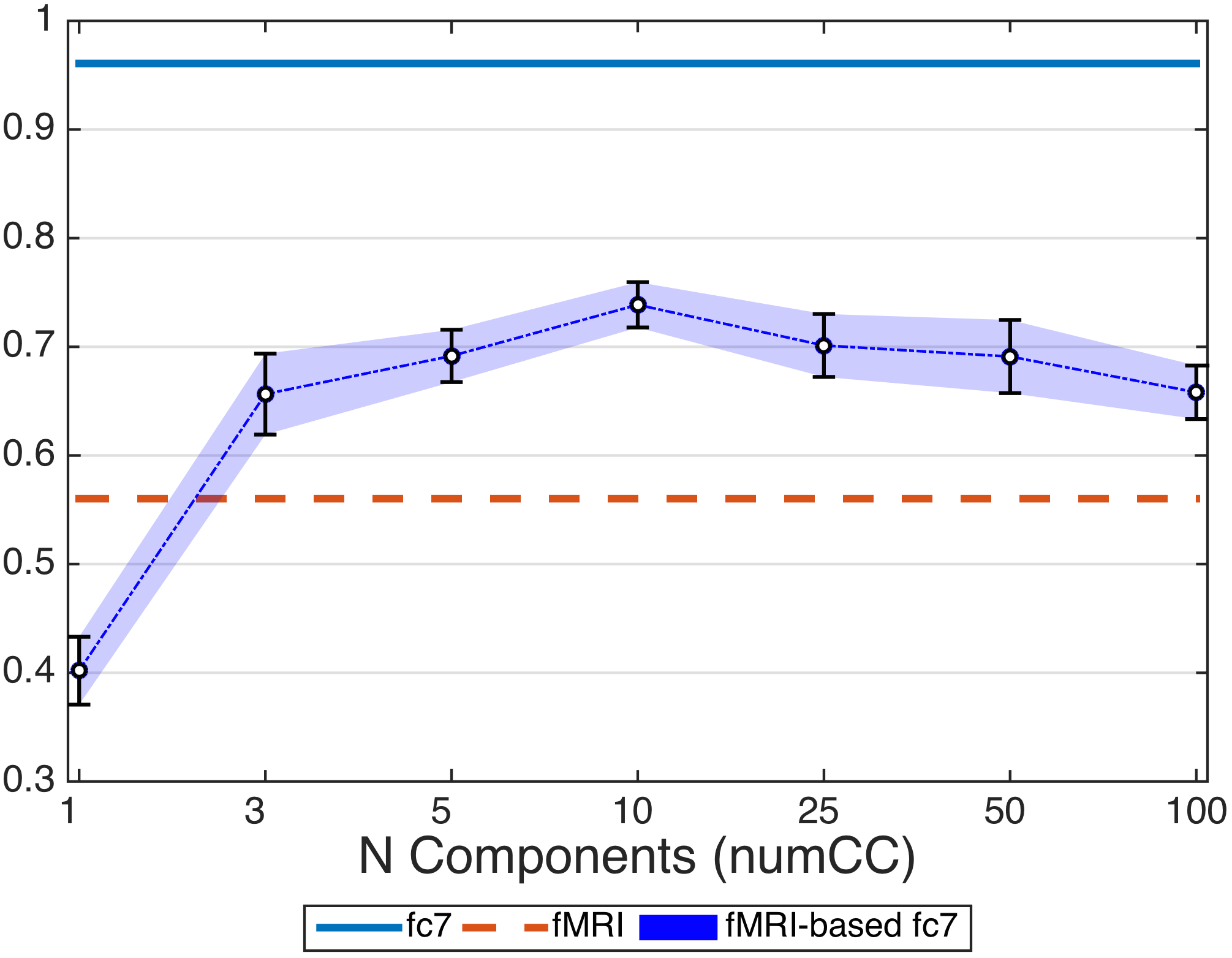}}
\subfigure[F1 measure]
	{\includegraphics[width=0.45\textwidth]{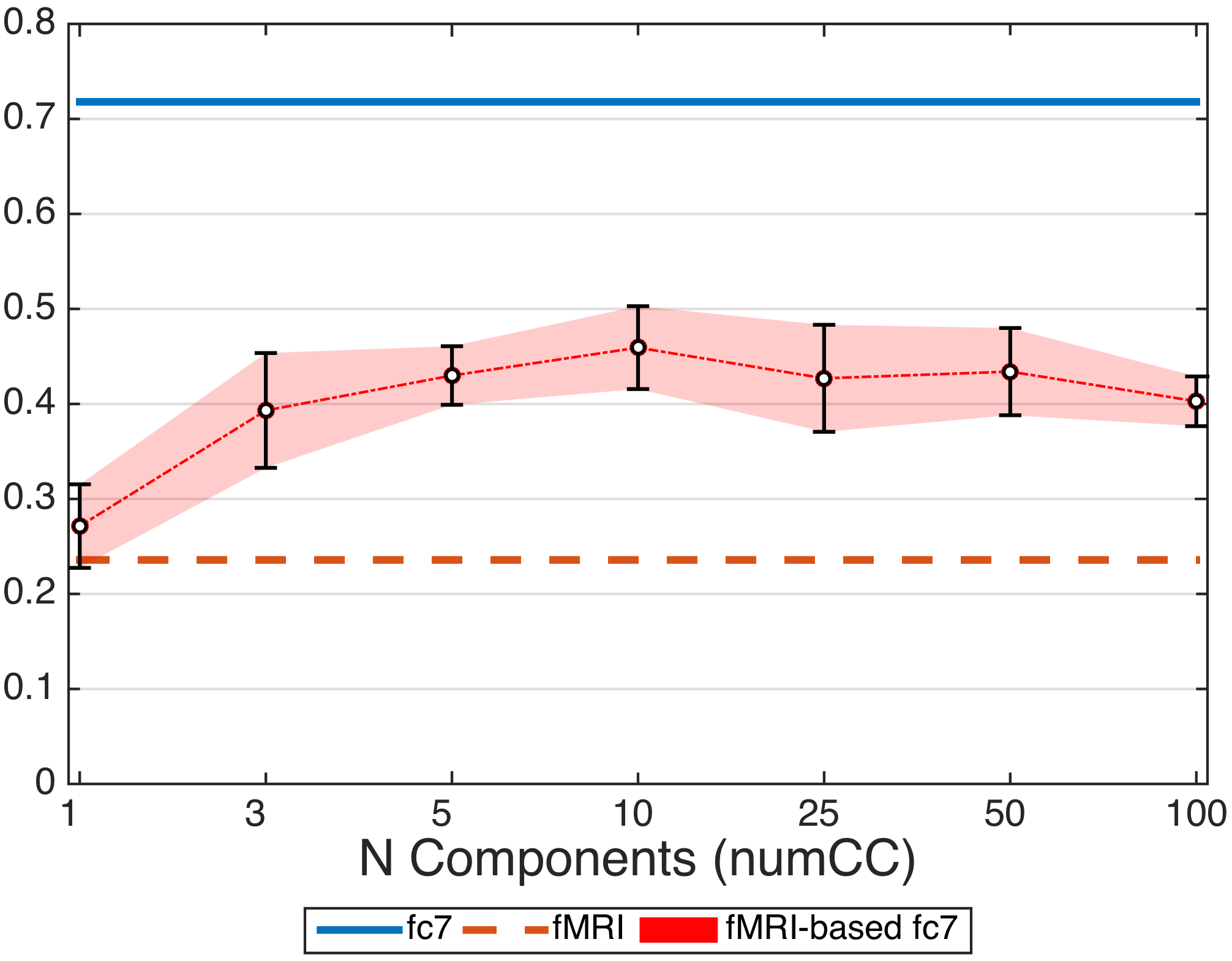}}
\caption{Classification performance comparison between: $fc7$ features only (blue), fMRI data only (red), and our method for predicting $fc7$ features from fMRI data across subjects. Performance are shown in terms of a) Accuracy and b) F1-measure. Shading describes standard deviation across subjects.}
\label{fig:classification}
\end{figure}

All three results (fMRI only, $fc7$ predicted from fMRI, and $fc7$ only) are presented in Figure~\ref{fig:classification}, in terms of a) \textit{accuracy} (i.e. the ratio between true results to the total number of examined cases), and b) \textit{F1-measure} (i.e. a weighted average of precision and recall).
As we expected, our method significantly improves the classification performance with respect to the classifier trained with fMRI data only, both in terms of accuracy (up to $55\%$) and F1-measure (up to $80\%$).
Best performance are obtained with 10 components, which is a sufficiently large number to exploit the discriminative properties of $fc7$ features, but small enough to keep good classification performance. Results also show a low variability across different subjects, thus underling once again the ability of the proposed method to generalize well across subjects.

\section{Conclusion and Outlook}
\label{sec:conclusions}

The present work describes an empirical study linking fMRI data taken while watching movies and deep neural network video representation. Excellent results in terms of correlation are obtained across different subjects and good results across different movies.
Preliminary results are also shown in a simple classification task, underling the ability of the method to effectively embed the imaging data onto a subspace more directly related to the classification task at hand. 
This could have a large impact on the neuroscientific studies where this embedding steps could facilitate fine-grained classification tasks.
In the future, we aim to extend the present work in several directions, among which broadening the analysis to other classes (car, house, dog, etc.) and other movies. Furthermore, it is interesting to examine the effects of incorporating different layers of CNNs.

\bibliographystyle{IEEEtran} 	

\begin{thebibliography}{10}
\providecommand{\url}[1]{#1}
\csname url@samestyle\endcsname
\providecommand{\newblock}{\relax}
\providecommand{\bibinfo}[2]{#2}
\providecommand{\BIBentrySTDinterwordspacing}{\spaceskip=0pt\relax}
\providecommand{\BIBentryALTinterwordstretchfactor}{4}
\providecommand{\BIBentryALTinterwordspacing}{\spaceskip=\fontdimen2\font plus
\BIBentryALTinterwordstretchfactor\fontdimen3\font minus
  \fontdimen4\font\relax}
\providecommand{\BIBforeignlanguage}[2]{{%
\expandafter\ifx\csname l@#1\endcsname\relax
\typeout{** WARNING: IEEEtran.bst: No hyphenation pattern has been}%
\typeout{** loaded for the language `#1'. Using the pattern for}%
\typeout{** the default language instead.}%
\else
\language=\csname l@#1\endcsname
\fi
#2}}
\providecommand{\BIBdecl}{\relax}
\BIBdecl

\bibitem{NKN11}
T.~Naselaris, K.~N. Kay, S.~Nishimoto, and J.~L. Gallant, ``Encoding and
  decoding in f{MRI},'' \emph{NeuroImage}, vol.~56, no.~2, pp. 400 -- 410,
  2011, multivariate Decoding and Brain Reading.

\bibitem{VME11}
G.~Valente, F.~D. Martino, F.~Esposito, R.~Goebel, and E.~Formisano,
  ``Predicting subject-driven actions and sensory experience in a virtual world
  with relevance vector machine regression of f{MRI} data,'' \emph{NeuroImage},
  vol.~56, no.~2, pp. 651 -- 661, 2011, multivariate Decoding and Brain
  Reading.

\bibitem{RSC17}
R.~G., M.~Svanera, G.~Gilam, M.~B. Cohen, T.~Lin, R.~Admon, T.~Gonen,
  A.~Thaler, R.~Goebel, S.~Benini, and G.~Valente, ``Robust inter-subject
  audiovisual decoding in fmri using kernel ridge regression,''
  \emph{Proceedings of the National Academy of Sciences (PNAS, under
  revision)}, 2017.

\bibitem{KNP08}
K.~N. Kay, T.~Naselaris, R.~J. Prenger, and J.~L. Gallant, ``Identifying
  natural images from human brain activity,'' \emph{Nature}, vol. 452, no.
  7185, pp. 352--355, 2008.

\bibitem{HMB07}
D.~R. Hardoon, J.~Mourão-Miranda, M.~Brammer, and J.~Shawe-Taylor,
  ``Unsupervised analysis of f{MRI} data using kernel canonical correlation,''
  \emph{NeuroImage}, vol.~37, no.~4, pp. 1250 -- 1259, 2007.

\bibitem{BSB11}
M.~B. Blaschko, J.~A. Shelton, A.~Bartels, C.~H. Lampert, and A.~Gretton,
  ``Semi-supervised kernel canonical correlation analysis with application to
  human f{MRI},'' \emph{Pattern Recognition Letters}, vol.~32, no.~11, pp. 1572
  -- 1583, 2011.

\bibitem{HDL10}
X.~Hu, F.~Deng, K.~Li, T.~Zhang, H.~Chen, X.~Jiang, J.~Lv, D.~Zhu, C.~Faraco,
  D.~Zhang \emph{et~al.}, ``Bridging low-level features and high-level
  semantics via f{MRI} brain imaging for video classification,'' in
  \emph{Proceedings of the international conference on Multimedia}.\hskip 1em
  plus 0.5em minus 0.4em\relax ACM, 2010, pp. 451--460.

\bibitem{JZH12}
X.~Jiang, T.~Zhang, X.~Hu, L.~Lu, J.~Han, L.~Guo, and T.~Liu, ``Music/speech
  classification using high-level features derived from fmri brain imaging,''
  in \emph{Proceedings of the 20th ACM international conference on
  Multimedia}.\hskip 1em plus 0.5em minus 0.4em\relax ACM, 2012, pp. 825--828.

\bibitem{TCT15}
V.~Tsatsishvili, F.~Cong, P.~Toiviainen, and T.~Ristaniemi, ``Combining pca and
  multiset cca for dimension reduction when group ica is applied to decompose
  naturalistic fmri data,'' in \emph{2015 International Joint Conference on
  Neural Networks (IJCNN)}.\hskip 1em plus 0.5em minus 0.4em\relax IEEE, 2015,
  pp. 1--6.

\bibitem{GvG15}
U.~G{\"u}{\c c}l{\"u} and M.~A.~J. van Gerven, ``Deep neural networks reveal a
  gradient in the complexity of neural representations across the ventral
  stream,'' \emph{The Journal of Neuroscience}, vol.~35, no.~27, pp.
  10\,005--10\,014, 2015.

\bibitem{ASM14}
P.~Agrawal, D.~Stansbury, J.~Malik, and J.~L. Gallant, ``Pixels to voxels:
  modeling visual representation in the human brain,'' \emph{arXiv preprint
  arXiv:1407.5104}, 2014.

\bibitem{KSH12}
A.~Krizhevsky, I.~Sutskever, and G.~E. Hinton, ``Imagenet classification with
  deep convolutional neural networks,'' in \emph{Advances in Neural Information
  Processing Systems 25}.\hskip 1em plus 0.5em minus 0.4em\relax Curran
  Associates, Inc., 2012, pp. 1097--1105.

\bibitem{CHY14}
C.~F. Cadieu, H.~Hong, D.~L. Yamins, N.~Pinto, D.~Ardila, E.~A. Solomon, N.~J.
  Majaj, and J.~J. DiCarlo, ``Deep neural networks rival the representation of
  primate it cortex for core visual object recognition,'' \emph{PLoS Comput
  Biol}, vol.~10, no.~12, p. e1003963, 2014.

\bibitem{KK14}
S.-M. Khaligh-Razavi and N.~Kriegeskorte, ``Deep supervised, but not
  unsupervised, models may explain it cortical representation,'' \emph{PLoS
  Comput Biol}, vol.~10, no.~11, p. e1003915, 2014.

\bibitem{CKP16}
R.~M. Cichy, A.~Khosla, D.~Pantazis, A.~Torralba, and A.~Oliva, ``Comparison of
  deep neural networks to spatio-temporal cortical dynamics of human visual
  object recognition reveals hierarchical correspondence,'' \emph{Scientific
  reports}, vol.~6, 2016.

\bibitem{YD16}
D.~L. Yamins and J.~J. DiCarlo, ``Using goal-driven deep learning models to
  understand sensory cortex,'' \emph{Nature neuroscience}, vol.~19, no.~3, pp.
  356--365, 2016.

\bibitem{RGH16}
S.~Ren, K.~He, R.~Girshick, and J.~Sun, ``Faster {R}-{CNN}: Towards real-time
  object detection with region proposal networks,'' in \emph{Advances in Neural
  Information Processing Systems}, 2015, pp. 91--99.

\bibitem{DJV14}
J.~Donahue, Y.~Jia, O.~Vinyals, J.~Hoffman, N.~Zhang, E.~Tzeng, and T.~Darrell,
  ``{DeCAF: A Deep Convolutional Activation Feature for Generic Visual
  Recognition},'' \emph{Icml}, vol.~32, pp. 647--655, 2014.

\bibitem{HZR15}
K.~He, X.~Zhang, S.~Ren, and J.~Sun, ``Deep residual learning for image
  recognition,'' \emph{CoRR}, vol. abs/1512.03385, 2015.

\bibitem{B36}
H.~Hotelling, ``Relations between two sets of variates,'' \emph{Biometrika},
  vol.~28, no. 3/4, pp. 321--377, 1936.

\bibitem{RTW16}
G.~Raz, A.~Touroutoglou, C.~Wilson-Mendenhall, G.~Gilam, T.~Lin, T.~Gonen,
  Y.~Jacob, S.~Atzil, R.~Admon, M.~Bleich-Cohen, A.~Maron-Katz, T.~Hendler, and
  L.~F. Barrett, ``Functional connectivity dynamics during film viewing reveal
  common networks for different emotional experiences,'' \emph{Cognitive,
  Affective, {\&} Behavioral Neuroscience}, pp. 1--15, 2016.

\bibitem{ICBM}
\BIBentryALTinterwordspacing
``{ICBM} atlases.'' [Online]. Available: \url{http://www.loni.usc.edu/atlases}
\BIBentrySTDinterwordspacing

\bibitem{BSA16}
S.~Benini, M.~Svanera, N.~Adami, R.~Leonardi, and A.~B. Kov{\'a}cs, ``Shot
  scale distribution in art films,'' \emph{Multimedia Tools and Applications},
  pp. 1--29, 2016.

\bibitem{BG15}
N.~Y. Bilenko and J.~L. Gallant, ``Pyrcca: regularized kernel canonical
  correlation analysis in python and its applications to neuroimaging,''
  \emph{arXiv preprint arXiv:1503.01538}, 2015.

\end{thebibliography}


\end{document}